\documentclass{article} 
\usepackage{colm}

\usepackage[
    a4paper,
    left=2.5cm,
    right=2.5cm,
    top=2.5cm,
    bottom=2.5cm
]{geometry}

\usepackage{microtype}
\usepackage{amsmath,amssymb,bm,mathtools}
\usepackage{xcolor}
\usepackage{hyperref}
\usepackage{booktabs}
\usepackage{graphicx}
\usepackage{placeins}
\AddToHook{env/table/begin}{\setlength{\belowcaptionskip}{7pt}}
\AddToHook{env/table*/begin}{\setlength{\belowcaptionskip}{7pt}}
\definecolor{darkblue}{rgb}{0,0,0.5}
\hypersetup{colorlinks=true,citecolor=darkblue,linkcolor=darkblue,urlcolor=darkblue}

\newcommand{\sg}{\operatorname{sg}}

\newcommand{\softplus}{\operatorname{softplus}}

\newcommand{\R}{\mathbb R}
\newcommand{\metricci}[3]{\begin{tabular}[t]{@{}c@{}}#1\\[-1pt]{\footnotesize[#2, #3]}\end{tabular}}

\title{A Health Trajectory Model with Language-Informed Data Assimilation}

\author{
    Songchi Zhou$^{1}$ \quad
    Sihang Zeng$^{2}$ \quad
    Tianle Hu$^{3}$ \quad
    Sheng Yu$^{1,*}$ \\[0.5em]
    \textnormal{$^{1}$ Department of Statistics and Data Science, Tsinghua University} \\
    \textnormal{$^{2}$ Department of Biomedical Informatics, Harvard Medical School} \\
    \textnormal{$^{3}$ Department of Computer Science and Technology, Tsinghua University} \\[0.3em]
    \textnormal{$^{*}$ Corresponding author:
    \texttt{syu@tsinghua.edu.cn}}
}

\colmfinalcopy 

\begin{document}

\maketitle

\begin{abstract}
Longitudinal electronic health records (EHRs) document a patient's hospital course through irregular measurements, coded events and clinical text. Individual measurements can be sparse, and a fixed panel of variables can omit evidence recorded elsewhere in the chart. We introduce HealthArc, a continuous-time trajectory model based on language-informed data assimilation. Between updates, a neural ordinary differential equation evolves the latent state, with a separate control vector encoding recorded interventions. At each update, a shared language encoder represents the available observations from their clinical descriptions and any recorded values and units. The difference between this embedding and the one predicted from the propagated state guides the correction, so even a single observation can update the trajectory without imputation. We trained HealthArc separately on MIMIC-IV and eICU, with 0.42 million training admissions across the two cohorts. On matched inputs, HealthArc reduced first-week physiological standardized mean absolute error by 6.7\% and 3.4\%, respectively, relative to the strongest baseline in each cohort. Uno's concordance for in-hospital mortality reached 0.814 and 0.747, compared with 0.808 and 0.657 for the strongest survival baselines. Using full structured histories with the same architecture raised concordance to 0.859 and 0.783. In both cohorts, the inferred states distinguished subsequent circulatory failure, severe respiratory failure and hyperglycemia. Unsupervised clustering identified groups with different clinical profiles and hospital survival. Latent-state neighbors had more similar measured organ function than neighbors selected by charting patterns. These results show that language-informed assimilation can integrate sparse clinical observations into trajectories that retain information about a patient's subsequent course.
\end{abstract}

\section{Introduction}
\label{sec:introduction}

During a hospital stay, a patient's evolving condition is documented piecemeal through numerical measurements, coded events and, in general clinical systems, free-text notes \citep{hripcsak2013next,rajkomar2018scalable}. Vital signs may be charted many times between successive blood tests \citep{cheng2020sparse}, leaving many individual variables sparse even as the overall chart grows and the subset of recorded measurements shifts across assessments \citep{debrouwer2019gru}. Observation timing also reflects clinical decisions: whether a laboratory test is ordered and how quickly it is repeated can carry prognostic signal alongside its value \citep{agniel2018biases}. Conventional clinical time-series models encode part of this sampling process by tracking missingness indicators or the time elapsed since each variable in a predefined panel was last measured \citep{harutyunyan2019multitask,che2018grud}. A model restricted to fixed variable channels, however, registers that a target laboratory value is absent yet overlooks relevant clinical evidence documented elsewhere in the chart. In an intensive care unit (ICU) mortality study, for example, adding clinical notes to physiological time series improved predictive performance \citep{khadanga2019notes}. Tracking a patient's condition therefore requires propagating earlier evidence across irregular gaps while incorporating new observations whenever they arrive in a different clinical form.

Recent longitudinal EHR models approach irregular histories along two lines. Generative event-sequence models forecast which clinical records will appear next, covering diagnoses and death in Delphi \citep{shmatko2025delphi} and extending to investigations, medications and associated numerical values in SurvivEHR \citep{gadd2026survivehr}. Once a blood pressure reading or laboratory test is charted, however, it also provides new evidence that should refine an ongoing estimate of the patient's condition. Continuous-time state-space models address this dual role. HealthFlux evolves a continuous latent health state trained to predict the next clinical event and its timing, revising that state recurrently from discrete clinical event embeddings or modality-specific encodings of prespecified numerical features \citep{wang2026healthflux}. Yet this architecture still depends on fixed numerical channels or separate modality encoders, preventing clinical entries recorded outside the predefined schema from directly refining the latent state. Assimilating heterogeneous, irregularly arriving clinical observations into a unified continuous-time trajectory without fixed panels or imputation remains an open challenge.

Natural clinical language offers a common format for diverse observations without committing to a fixed input schema. Pretrained language models can encode patient histories written as medical code descriptions, with values and units included when available, for downstream outcome prediction \citep{hegselmann2026encoders}. In parallel, predictive-representation models such as Clin-JEPA encode the observations available in each hour and forecast subsequent hourly latent representations conditioned on interventions \citep{yang2026clinjepa}. Combining a shared language interface with representation prediction suggests a direct way to assimilate irregular records: if a continuous latent state predicts the language embedding of the next clinical observation, the difference between the observed and predicted embeddings can guide a state update even when consecutive entries describe different clinical variables.

\begin{figure}[t]
    \centering
    \includegraphics[width=\linewidth]{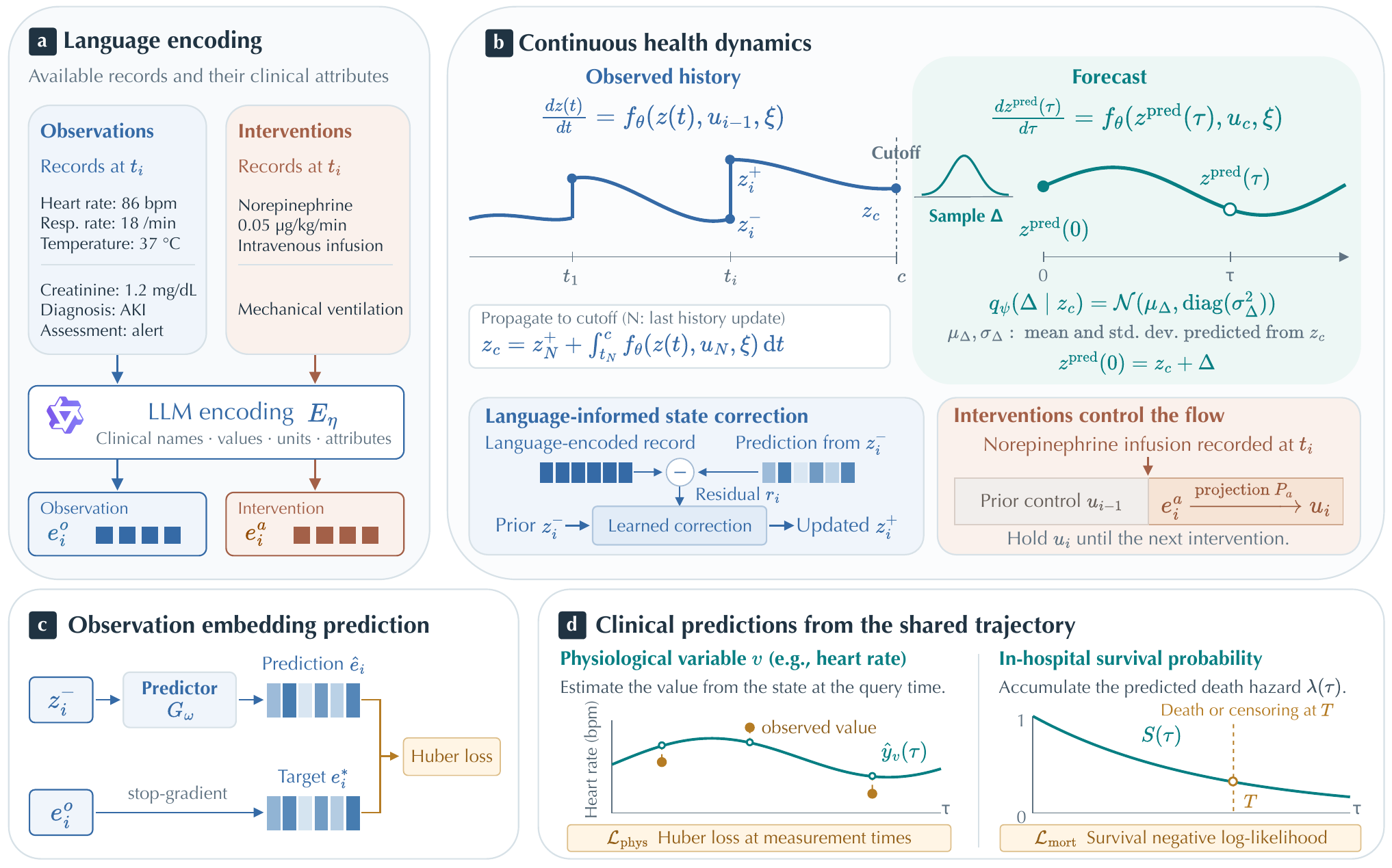}
    \caption{\textbf{HealthArc overview.} (a) At recording time $t_i$, a shared language encoder $E_\eta$ embeds available clinical observations into $e_i^o$ and recorded interventions separately into $e_i^a$. (b) A shared neural ODE vector field $f_\theta$ evolves the latent state continuously between updates and across the forecast horizon. At each observation update, the prediction residual $r_i$ corrects the propagated state ($z_i^- \to z_i^+$), while recorded interventions update the external control vector $u_i$ for subsequent evolution. From the cutoff state $z_c$, training samples a variational offset $\Delta$, whereas evaluation starts deterministic point forecasts from the posterior mean. (c) Before each observation update, a predictor $G_\omega$ anticipates the incoming observation embedding and is supervised by a Huber loss against the normalized target embedding with stop-gradient applied. (d) From the shared forecast trajectory $z^{\mathrm{pred}}(\tau)$, the physiological prediction head estimates a queried variable $v$ as $\hat y_v(\tau)$ at elapsed time $\tau$, and the mortality prediction head integrates the instantaneous hazard to obtain survival probability $S(\tau)$.}
    \label{fig:overview}
\end{figure}

Here we introduce HealthArc, a continuous-time trajectory model based on language-informed data assimilation (Figure~\ref{fig:overview}). HealthArc evolves a patient's latent health state continuously under a neural ordinary differential equation (ODE) conditioned on separately encoded clinical interventions, while updating that state whenever new observations arrive using the prediction residual between observed and anticipated language embeddings (Section~\ref{sec:method}). Through this shared language space, even a single measurement or bedside assessment can update the trajectory without fixed variable panels or missingness imputation. Across two large inpatient cohorts, MIMIC-IV and eICU, we evaluate HealthArc on physiological forecasting and in-hospital mortality prediction, comparing against task-specific baselines on matched clinical variables and testing expansion to full structured clinical histories. We also examine the clinical structure of the learned latent states and trajectories (Sections~\ref{sec:experiments}--\ref{sec:results}).

\section{Related work}
\label{sec:related-work}

\paragraph{Longitudinal EHR foundation models.}
Several foundation models represent a patient's longitudinal record as a sequence of discrete clinical events that can be extended forward in time, including Foresight and ETHOS \citep{kraljevic2024foresight,renc2024ethos}. Delphi generates candidate disease trajectories by iteratively sampling the next diagnosis and its waiting time conditioned on past history \citep{shmatko2025delphi}, while ReClaim models disease onset and healthcare expenditure from medical claims \citep{ma2026reclaim}. Other models forecast numerical values alongside clinical events. SurvivEHR predicts which event occurs next across the record, when it occurs and, for numerical investigations, its value \citep{gadd2026survivehr}. ORA instead models the time to the next occurrence of each individual code together with its associated value when available \citep{jing2026ora}. DT-GPT forecasts numerical clinical trajectories using language models \citep{makarov2025dtgpt}, whereas time-conditioned forecasting predicts the event type at a queried future time followed by its value token conditioned on that type \citep{kang2026tcf}. A complementary group of models learns representations directly for downstream clinical queries. APOLLO integrates structured events, clinical text and images for downstream prediction \citep{zhang2026apollo}, whereas MOTOR pretrains representations on right-censored time-to-event tasks \citep{steinberg2024motor}. More recently, ICareFM estimates the risk of crossing prespecified clinical thresholds \citep{burger2026icarefm}, and EveryQuery conditions predictions on structured task queries \citep{chandak2026everyquery}. Whereas these approaches focus on discrete event generation, time-to-event risk modeling or structured query interfaces, HealthArc assimilates heterogeneous clinical observations in a shared language space to maintain a single continuous-time latent health trajectory.

\paragraph{Continuous-time models and state estimation.}
State estimation seeks to infer a latent clinical condition from sparse, noisy measurements. In classical linear filtering, the residual between an incoming measurement and its model prediction drives the state correction \citep{kalman1960filtering}. Neural continuous-time models extend this idea to nonlinear, irregularly sampled series. ODE-RNNs and neural continuous-discrete state-space models propagate hidden states or belief distributions across irregular intervals \citep{rubanova2019latent,ansari2023ncdssm}. GRU-ODE-Bayes alternates continuous latent evolution with Bayesian-style updates at observed entries \citep{debrouwer2019gru}, whereas neural controlled differential equations (neural CDEs) drive latent dynamics along an interpolated path of observations \citep{kidger2020neural}. In clinical survival analysis, SurvLatent ODE models competing risks from continuous latent trajectories \citep{moon2022survlatent}, while TrajSurv aligns neural CDE trajectories with clinical severity scores through a contrastive objective \citep{zeng2025trajsurv}. HealthFlux couples continuous latent state evolution with recurrent observation updates across clinical and biological modalities \citep{wang2026healthflux}. HealthArc builds on this continuous-discrete paradigm by computing state corrections directly from the discrepancy between predicted and observed language representations of clinical entries.

\paragraph{Predictive representations and observation updates.}
Predicting features of upcoming observations provides a reference for interpreting new data. Predictive-state inference machines learn filtering updates by predicting future observation features \citep{sun2016psim}, while joint-embedding predictive architectures learn to predict target representations in latent embedding space rather than raw input space \citep{assran2023ijepa}. Within longitudinal EHRs, SMB-Structure combines next-token prediction with an embedding prediction objective over partially masked future records \citep{adam2026smb}, and Clin-JEPA forecasts future hourly state representations conditioned on clinical actions \citep{yang2026clinjepa}. HealthArc applies representation prediction at each irregular update time: the propagated latent state anticipates the language embedding of the incoming observations, and the difference between the actual and predicted embeddings guides each state revision.

\section{HealthArc: Continuous-time dynamics and language-informed assimilation}
\label{sec:method}

Figure~\ref{fig:overview} summarizes the HealthArc architecture. Given an irregular clinical history of static baseline attributes, clinical observations and interventions (Section~\ref{sec:task}), we map the clinical entries recorded at each update into a shared language space and assimilate the pre-cutoff history by alternating continuous-time neural ODE propagation with residual-based observation updates, while recorded interventions update an external control vector and a step-wise embedding prediction loss supervises the pre-update predictor (Sections~\ref{sec:latent-filter}--\ref{sec:historical-semantics}). At the forecast cutoff, we introduce a variational initial-state offset and evolve the shared latent trajectory forward under the same vector field to drive physiological and mortality prediction heads trained jointly (Sections~\ref{sec:forecast-dynamics}--\ref{sec:objective}).

\subsection{Problem formulation and clinical record structure}
\label{sec:task}

For a hospital admission $p$, let $t_0 = 0$ denote the time of admission and let $c > 0$ denote the forecast cutoff measured in hours since admission. Let $\mathcal S_p$ denote any static baseline information available for the admission, such as sex, birth date or ethnicity, which may be empty when inputs are restricted to timestamped clinical variables. Up to the cutoff $c$, the observed clinical history is
\begin{equation}
    \mathcal H_{p,c}
    = \left(\mathcal S_p,\,
    \left\{(t_{p,i}, \mathcal O_{p,i}, \mathcal A_{p,i})\right\}_{i=1}^{N_p}\right),
    \qquad 0 \le t_{p,1} < \cdots < t_{p,N_p} \le c,
    \label{eq:history}
\end{equation}
where $t_{p,i}$ is the $i$-th recording time in hours since admission, $N_p$ is the number of retained recording times up to $c$, $\mathcal O_{p,i}$ is the set of clinical \emph{observations} recorded at $t_{p,i}$, and $\mathcal A_{p,i}$ is the set of clinical \emph{interventions} recorded at $t_{p,i}$. Either $\mathcal O_{p,i}$ or $\mathcal A_{p,i}$ may be empty at a given timestamp.

\paragraph{Observations versus interventions.}
Observations $\mathcal O_{p,i}$ are clinical measurements and assessments that reflect the patient's current condition, including vital signs, numerical or qualitative laboratory results, bedside assessments, fluid outputs, care-plan items and documented diagnoses. Interventions $\mathcal A_{p,i}$ are recorded therapeutic actions and orders intended to alter the patient's subsequent course, such as documented medication orders or administrations, intravenous infusions, mechanical ventilation, renal replacement therapy and procedures. Separating these two streams follows the state-space distinction between state measurements and external control inputs: observations serve as measurement evidence that updates the current latent state through prediction residuals, whereas interventions are excluded from observation prediction and instead update a control vector that conditions subsequent continuous-time evolution. When an intervention is followed by a rapid physiological change, subsequent observations register that shift and revise the latent state accordingly. In structured electronic health records, clinical entries can be routed into $\mathcal O_{p,i}$ or $\mathcal A_{p,i}$ according to standardized event categories (Section~\ref{sec:full-variable} and Appendix~\ref{app:language-interface}). Extending this partition to unstructured narrative clinical notes is discussed in Section~\ref{sec:limitations}.

\paragraph{Forecasting objective.}
Given $\mathcal H_{p,c}$, we aim to forecast the patient's clinical trajectory over a future horizon $H > 0$ measured in hours after the cutoff. Let $\tau \in (0, H]$ denote elapsed time since the cutoff $c$, corresponding to time $c + \tau$ since admission. For any $\tau \in (0, H]$, we target two clinical predictions: the value $\hat y_v(\tau)$ of a queried physiological variable $v$ at time $\tau$, and the estimated net mortality risk $1 - S(\tau)$ over $(0, \tau]$, where $S(\tau)$ denotes the survival probability through elapsed time $\tau$ under right censoring of live discharges (Section~\ref{sec:objective}). For readability, we omit the admission index $p$ in the remainder of Section~\ref{sec:method}.

\subsection{Language encoding, continuous dynamics and residual-based updates}
\label{sec:latent-filter}

During the observed history ($t \in [0, c]$), we maintain a continuous-time latent health state $z(t) \in \R^{d_z}$ of dimension $d_z$ alongside an intervention control vector $u_i \in \R^{d_u}$ of dimension $d_u$ that stays constant between updates. At each pre-cutoff update time $t_i$ ($i = 1, \dots, N$), we denote the state propagated to $t_i$ just before assimilating observations by $z_i^-$ and the updated state afterward by $z_i^+$. Let $m_i^o = \mathbb I(\mathcal O_i \neq \varnothing)$ and $m_i^a = \mathbb I(\mathcal A_i \neq \varnothing)$ indicate whether observations and interventions are present at $t_i$.

\paragraph{Language encoding and initialization.}
A shared pretrained language encoder $E_\eta$ maps serialized clinical text into a common embedding space $\R^{d_e}$ of dimension $d_e$, allowing variable combinations of clinical entries to enter the model without fixed panels or imputation. Whenever observations or interventions are present at $t_i$, we serialize each set into plain-text clinical descriptions (including any recorded numerical or textual values and units) and apply $E_\eta$. Static baseline attributes $\mathcal S$ are likewise encoded via $E_\eta$ into a pooled static input $e_{\mathrm{stat}}$ and mapped by a learned projection $P_{\mathrm{stat}}$ into a static context vector $\xi \in \R^{d_\xi}$ of dimension $d_\xi$:
\begin{equation}
    e_i^o = E_\eta(\mathcal O_i) \in \R^{d_e}\;\;(m_i^o = 1),
    \qquad
    e_i^a = E_\eta(\mathcal A_i) \in \R^{d_e}\;\;(m_i^a = 1),
    \qquad
    \xi = P_{\mathrm{stat}}(e_{\mathrm{stat}}) \in \R^{d_\xi},
    \label{eq:language-encoding}
\end{equation}
where $E_\eta(\cdot)$ operates on the serialized text of each set, and $e_{\mathrm{stat}}$ is zero-filled when $\mathcal S = \varnothing$ so that $P_{\mathrm{stat}}$ outputs a shared default static context (Appendix~\ref{app:language-interface}). At admission ($t_0 = 0$), we initialize the latent state and control vector as
\begin{equation}
    z_0^+ = z_{\mathrm{init}} + P_{\mathrm{init}}(\xi),
    \qquad
    u_0 = \mathbf 0 \in \R^{d_u},
    \label{eq:initialization}
\end{equation}
where $z_{\mathrm{init}} \in \R^{d_z}$ is a learned initial state vector and $P_{\mathrm{init}}$ is a learned projection of $\xi$.

\paragraph{Intervention control sequence.}
Interventions act as external control inputs to the latent dynamics rather than state measurements, updating the control sequence via a learned projection network $P_a$:
\begin{equation}
    u_i =
    \begin{cases}
        P_a(e_i^a), & m_i^a = 1,\\
        u_{i-1}, & m_i^a = 0,
    \end{cases}
    \qquad i = 1, \dots, N.
    \label{eq:control-update}
\end{equation}
An intervention recorded at $t_i$ updates $u_i$ to govern the subsequent continuous-time evolution without inducing an instantaneous jump in $z(t)$. When no intervention is recorded at $t_i$ ($m_i^a = 0$), the active control $u_{i-1}$ carries forward, and admissions without recorded interventions run unchanged with $u_i = \mathbf 0$ throughout.

\paragraph{Continuous-time state propagation.}
Between successive update times $t_{i-1}$ and $t_i$ ($i = 1, \dots, N$), a neural network vector field $f_\theta$ governs the continuous-time evolution of $z(t)$ with the preceding active control $u_{i-1}$ and static context $\xi$ held fixed:
\begin{equation}
    \frac{d z(t)}{d t} = f_\theta\!\left(z(t), u_{i-1}, \xi\right),
    \qquad t \in [t_{i-1}, t_i),
    \qquad z(t_{i-1}) = z_{i-1}^+.
    \label{eq:history-ode}
\end{equation}
Integrating Equation~\eqref{eq:history-ode} across $[t_{i-1}, t_i]$ yields the pre-update state at $t_i$ (with $z_1^- = z_0^+$ if $t_1 = 0$):
\begin{equation}
    z_i^- = z_{i-1}^+ + \int_{t_{i-1}}^{t_i} f_\theta\!\left(z(t), u_{i-1}, \xi\right)\,d t.
    \label{eq:prior-prediction}
\end{equation}

\paragraph{Residual-based observation update.}
Following continuous-discrete filtering \citep{kalman1960filtering,debrouwer2019gru,ansari2023ncdssm}, we update the latent state using the prediction residual of an incoming observation rather than overwriting it with raw inputs. Before assimilating $\mathcal O_i$, a neural predictor $G_\omega$ maps the propagated state $z_i^-$ into the shared language space, producing $G_\omega(z_i^-) \in \R^{d_e}$ as the observation embedding predicted from the history up to $t_{i-1}$. Whenever observations are present ($m_i^o = 1$), we apply the same normalization $\mathrm{LN}_a$ to the observed and predicted embeddings and subtract the prediction from the observation to obtain the residual $r_i \in \R^{d_e}$, which drives the state correction (while $z_i^+ = z_i^-$ when $m_i^o = 0$):
\begin{equation}
    \begin{aligned}
        r_i &= \mathrm{LN}_a(e_i^o) - \mathrm{LN}_a\!\left(G_\omega(z_i^-)\right),\\
        z_i^+ &= z_i^- + C_\phi(z_i^-, r_i),\\
        C_\phi(z, r) &= F_\phi\!\left([\mathrm{LN}_z(z),\, r]\right) \odot \tanh(W_\phi r).
    \end{aligned}
    \label{eq:observation-update}
\end{equation}
Here $\mathrm{LN}_a$ is a Layer Normalization layer with learned coordinate-wise scale and bias applied identically to $e_i^o$ and $G_\omega(z_i^-)$ before subtraction, so the shared bias cancels while the learned scale weights each coordinate of the standardized embedding difference. Separately, $\mathrm{LN}_z$ normalizes $z_i^-$ with its own learned scale and bias before concatenation $[\cdot, \cdot]$ with $r_i$, allowing updates to depend on the patient's current state. The correction module $C_\phi$ pairs a neural network $F_\phi$ (outputting a candidate correction in $\R^{d_z}$) with a bias-free linear projection $W_\phi \in \R^{d_z \times d_e}$ and elementwise multiplication $\odot$, applying signed coordinate-wise gating via $\tanh(W_\phi r_i)$. When the normalized embeddings match, $r_i = \mathbf 0$ and the bias-free gate makes the correction zero, giving $z_i^+ = z_i^-$.

\paragraph{Timestamp-wise versus temporally grouped assimilation.}
Assimilating every retained recording time in Equation~\eqref{eq:history} directly defines \emph{timestamp-wise assimilation}, where the number of updates equals the number of recording times ($N = N_p$). In densely charted ICU admissions, \emph{temporally grouped assimilation} reduces update frequency without discarding clinical entries. Consecutive observation-only timestamps are concatenated in chronological order, each tagged with its time offset relative to the group's final timestamp, as long as the group spans at most a preset maximum window duration and fits within the encoder's context window. Each group then serves as a single observation set $\mathcal O_i$ encoded into $e_i^o$ and assimilated at its final timestamp $t_i$, reducing the update count to $N \le N_p$ while leaving Equations~\eqref{eq:language-encoding}--\eqref{eq:observation-update} unchanged. Any recorded intervention closes the preceding observation group and is processed at its exact timestamp together with any simultaneous observations (Appendix~\ref{app:language-interface}).

\subsection{Historical observation-embedding prediction}
\label{sec:historical-semantics}

The auxiliary loss trains $G_\omega$ to anticipate the embedding of incoming clinical observations from the pre-update state $z_i^-$ before the entries at $t_i$ are assimilated. When those observations arrive, Equation~\eqref{eq:observation-update} uses the difference between their normalized language embedding and the normalized prediction to revise the state.

Following predictive-representation learning \citep{assran2023ijepa,yang2026clinjepa}, we supervise $G_\omega(z_i^-)$ directly against $e_i^o$ at every observation update $t_i$. Unlike $\mathrm{LN}_a$ in Equation~\eqref{eq:observation-update}, this loss applies \emph{parameter-free} Layer Normalization $\mathrm{LN}_0$ (zero mean and unit variance per vector, without learned scale or bias) to form $\hat e_i = \mathrm{LN}_0(G_\omega(z_i^-))$ and $e_i^* = \sg[\mathrm{LN}_0(e_i^o)]$ (Figure~\ref{fig:overview}c). If $\mathrm{LN}_a$ were used inside the loss, the optimizer could trivially shrink the loss by scaling down its learned coordinate weights without improving $G_\omega$. Using parameter-free $\mathrm{LN}_0$ removes that shortcut, while the stop-gradient operator $\sg[\cdot]$ blocks gradients through $e_i^*$ so this auxiliary loss updates the predictor and trajectory dynamics toward the encoded observation rather than pulling the target representation directly.

At observation update $i$, averaging the Huber penalty $\rho$ across the $d_e$ coordinates of $\hat e_i - e_i^*$ gives the per-update loss $\ell_{{\rm hist}, i} = \frac{1}{d_e}\sum_{j=1}^{d_e} \rho(\hat e_{i,j} - e^*_{i,j})$. Letting $\mathcal I = \{i \mid m_i^o = 1\}$ index an admission's observation updates, the per-admission historical embedding prediction loss is
\begin{equation}
  \bar\ell_{\rm hist}
  = \frac{1}{|\mathcal I|}\sum_{i \in \mathcal I} \ell_{{\rm hist}, i}
  = \frac{1}{|\mathcal I|}\sum_{i \in \mathcal I}
    \frac{1}{d_e}\sum_{j=1}^{d_e} \rho\!\left(\hat e_{i,j} - e^*_{i,j}\right).
  \label{eq:historical-prediction-loss}
\end{equation}
Averaging $\bar\ell_{\rm hist}$ across admissions with at least one observation update yields the cohort objective $\mathcal L_{\rm hist}$ (Appendix~\ref{app:implementation}).

\subsection{Forecasting beyond the cutoff}
\label{sec:forecast-dynamics}

After the final pre-cutoff update at $t_N \le c$, we integrate the same vector field $f_\theta$ from $t_N$ to the forecast cutoff $c$ under the active control $u_c = u_N$ (with $z_c = z_N^+$ if $t_N = c$):
\begin{equation}
    z_c = z_N^+ + \int_{t_N}^{c} f_\theta\!\left(z(t), u_N, \xi\right)\,d t.
    \label{eq:cutoff-state}
\end{equation}
Since pre-cutoff records are sparse and partial, multiple underlying physiological states can be consistent with the same observed history. Following variational Latent ODEs \citep{rubanova2019latent,moon2022survlatent}, we model initial-state uncertainty at the cutoff by adding a latent offset $\Delta \in \R^{d_z}$ drawn from a diagonal Gaussian approximate posterior:
\begin{equation}
    q_\psi(\Delta \mid z_c) = \mathcal N\!\left(\mu_\Delta(z_c),\, \operatorname{diag}(\sigma_\Delta^2(z_c))\right),
    \label{eq:cutoff-posterior}
\end{equation}
where neural networks $\mu_\Delta$ and $\sigma_\Delta$ (parameterized by $\psi$) output the mean shift and coordinate-wise standard deviation. During training, a Kullback--Leibler penalty regularizes $q_\psi(\Delta \mid z_c)$ toward a zero-mean isotropic Gaussian prior $p(\Delta) = \mathcal N(\mathbf 0, \sigma_0^2 I_{d_z})$, where $\sigma_0 > 0$ is a preset prior-scale hyperparameter (Appendix~\ref{app:implementation}). This zero-centered prior regularizes the initial forecast state toward $z_c$, while $\mu_\Delta(z_c)$ allows a learned, history-dependent shift in its mean.

To distinguish post-cutoff trajectories from $z(t)$, we denote the forecast latent state at elapsed time $\tau \in [0, H]$ by $z^{\mathrm{pred}}(\tau)$, evolving under the same vector field $f_\theta$ with $u_c$ and $\xi$ held fixed:
\begin{equation}
    z^{\mathrm{pred}}(0) = z_c + \Delta,
    \qquad
    \frac{d z^{\mathrm{pred}}(\tau)}{d \tau} = f_\theta\!\left(z^{\mathrm{pred}}(\tau), u_c, \xi\right),
    \qquad 0 \le \tau \le H.
    \label{eq:cutoff-latent}
\end{equation}
During training, we draw one sample $\Delta \sim q_\psi(\Delta \mid z_c)$ per admission via the reparameterization trick and share $z^{\mathrm{pred}}(\tau)$ across both clinical prediction heads. At evaluation time, deterministic point forecasts start from the posterior mean $z^{\mathrm{pred}}(0) = z_c + \mu_\Delta(z_c)$.

\subsection{Clinical prediction heads and joint training objective}
\label{sec:objective}

\paragraph{Point-in-time physiological forecasting.}
To predict a physiological variable $v$ at elapsed time $\tau \in (0, H]$ on a standardized scale (Appendix~\ref{app:implementation}), let $m_v \in \{0, 1\}$ indicate whether $v$ was measured before the cutoff and let $b_v$ denote its last pre-cutoff standardized measurement (set to zero when $m_v = 0$). From $z^{\mathrm{pred}}(\tau)$, $v$, $\tau$, $m_v$ and $b_v$, the physiological prediction head $D_{\rm phys}$ outputs $\tilde y_v(\tau) = D_{\rm phys}(z^{\mathrm{pred}}(\tau), v, \tau, m_v, b_v)$ and forms the prediction
\begin{equation}
    \hat y_v(\tau) =
    \begin{cases}
        b_v + \bigl[1 - \exp\{-\softplus(\alpha_v)\tau\}\bigr]\,\tilde y_v(\tau), & m_v = 1,\\
        \tilde y_v(\tau), & m_v = 0,
    \end{cases}
    \label{eq:phys-anchor}
\end{equation}
where $\softplus(x) = \log(1 + e^x)$ and $\alpha_v$ is a learned scalar parameter specific to variable $v$. When $v$ was measured before the cutoff ($m_v = 1$), the time-dependent gate rises from zero at $\tau = 0$ toward one as $\tau$ grows, anchoring short-horizon forecasts to $b_v$, whereas unobserved variables ($m_v = 0$) are predicted directly. Let $\mathcal M = \{(v_k, \tau_k, y_k)\}_{k=1}^{M}$ denote the future standardized physiological measurements observed for an admission over $(0, H]$, with per-measurement error $\ell_{{\rm phys}, k} = \rho(\hat y_{v_k}(\tau_k) - y_k)$. Its per-admission physiological forecasting loss is
\begin{equation}
    \bar\ell_{\rm phys} = \frac{1}{M}\sum_{k=1}^{M} \ell_{{\rm phys}, k} = \frac{1}{M}\sum_{k=1}^{M} \rho\!\left(\hat y_{v_k}(\tau_k) - y_k\right),
    \label{eq:phys-loss}
\end{equation}
and averaging $\bar\ell_{\rm phys}$ across admissions with at least one future measurement yields $\mathcal L_{\rm phys}$.

\paragraph{Interval mortality risk.}
Following continuous-time survival ODEs \citep{tang2022soden}, the mortality prediction head $D_{\rm mort}$ maps $z^{\mathrm{pred}}(\tau)$ to a strictly positive instantaneous hazard of in-hospital death, $\lambda(\tau) = D_{\rm mort}(z^{\mathrm{pred}}(\tau)) > 0$, and the cumulative hazard evolves as $d \Lambda(\tau)/d \tau = \lambda(\tau)$ with $\Lambda(0) = 0$. We integrate $z^{\mathrm{pred}}(\tau)$ and $\Lambda(\tau)$ jointly as a coupled ODE system over $[0, H]$. For post-cutoff follow-up duration $T \in (0, H]$ and death indicator $\delta \in \{0, 1\}$ (with $\delta = 1$ at in-hospital death and $\delta = 0$ when censored at live discharge or horizon $H$), the cumulative hazard, survival probability and exact-time negative log-likelihood contribution are
\begin{equation}
    \Lambda(T) = \int_0^T \lambda(\tau)\,d \tau,
    \qquad
    S(T) = \exp[-\Lambda(T)],
    \qquad
    \ell_{\rm surv} = \Lambda(T) - \delta\log\lambda(T).
    \label{eq:mortality-loss}
\end{equation}
Averaging $\ell_{\rm surv}$ across admissions with valid mortality follow-up gives $\mathcal L_{\rm mort}$ (Figure~\ref{fig:overview}d). Under conditionally independent censoring given the observed history, the estimated net mortality risk by elapsed time $\tau \in (0, H]$ is $1 - S(\tau)$.

\paragraph{Joint learning.}
Physiological measurements and mortality outcomes jointly supervise $z^{\mathrm{pred}}(\tau)$. To encourage smooth latent dynamics, we average the per-admission integrated squared velocity $\ell_{\rm kin} = \int_0^T \|d z^{\mathrm{pred}}(\tau)/d \tau\|_2^2\,d \tau$ across admissions with positive mortality follow-up to form $\mathcal L_{\rm kin}$. To regularize the variational cutoff posterior, we average the per-coordinate Kullback--Leibler divergence $\ell_{\rm cut} = \frac{1}{d_z}\operatorname{KL}\!\left(q_\psi(\Delta \mid z_c)\,\|\,\mathcal N(\mathbf 0, \sigma_0^2 I_{d_z})\right)$ across admissions to form $\mathcal L_{\rm cut}$. Combining all five cohort-averaged terms yields the joint objective:
\begin{equation}
    \mathcal L =
    \mathbb E_{\Delta \sim q_\psi(\cdot \mid z_c)}\!\left[\mathcal L_{\rm phys}
      + \lambda_{\rm mort}\mathcal L_{\rm mort}
      + \lambda_{\rm kin}\mathcal L_{\rm kin}\right]
      + \lambda_{\rm hist}\mathcal L_{\rm hist}
      + \beta_{\rm cut}\mathcal L_{\rm cut},
    \label{eq:total-loss}
\end{equation}
where the expectation is approximated using one reparameterized sample $\Delta \sim q_\psi(\Delta \mid z_c)$ per admission, and $\lambda_{\rm mort}$, $\lambda_{\rm kin}$, $\lambda_{\rm hist}$ and $\beta_{\rm cut}$ are nonnegative weights relative to unit weight on $\mathcal L_{\rm phys}$. Under $\mathcal L$, we train all trajectory-model parameters and the low-rank adapters of the pretrained language encoder jointly in a single stage (Appendix~\ref{app:implementation}).

\section{Evaluation of health trajectories}
\label{sec:experiments}

We compare physiological and mortality forecasts with task-specific baselines using matched inputs in two inpatient cohorts. We also evaluate forecasts from full structured histories and measure the computational cost of grouping observations. Finally, we examine whether cutoff states distinguish subsequent acute events and how the learned trajectories relate to clinical course and hospital survival.

\subsection{Data and study cohorts}

MIMIC-IV v3.1 contains inpatient records from Beth Israel Deaconess Medical Center \citep{johnson2023mimic,johnson2024mimic31}, and the eICU Collaborative Research Database v2.0 contains critical care records from hospitals across the United States \citep{pollard2018eicu,pollard2019eicu20}. We obtained both datasets from PhysioNet \citep{pollard2026physionet} and used versions converted to the Medical Event Data Standard (MEDS) \citep{mcdermott2026meds}. Each MEDS conversion supplied an 80/10/10 split into training, tuning and test sets, partitioned by patient in MIMIC-IV and by hospital admission in eICU.

For each eligible admission, HealthArc observed the first $c = 36$ hours since admission and forecast 41 physiological variables spanning vital signs, serum chemistries, liver tests, blood counts and coagulation panels (Table~\ref{tab:channel-catalogue}), together with in-hospital mortality over the following week ($H = 168$\,h). Admissions were included if hospital discharge occurred after the 36-hour cutoff, total length of stay was at most 180 days, and at least eight historical entries were recorded by the cutoff. Table~\ref{tab:result-cohorts} summarizes the training and test cohorts together with the observations available for each test endpoint. To assess longer-range forecasting from the same 36-hour history, we also evaluated horizons at the 95th percentile of training follow-up ($H = 415$\,h in MIMIC-IV and $H = 494$\,h in eICU; Appendix~\ref{app:scoring}).

\begin{table}[!htb]
\centering
\small
\setlength{\tabcolsep}{3pt}
\caption{Training and test admissions in MIMIC-IV and eICU. Age and length of stay are reported as medians [interquartile ranges]. Test outcomes are shown at each forecast horizon: admissions with measured targets are test admissions with at least one post-cutoff measurement among the 41 physiological variables, and admissions in mortality comparison are test admissions with valid survival follow-up shared across all mortality models.}
\label{tab:result-cohorts}
\begin{tabular}{@{}lrrrr@{}}
\toprule
 & \multicolumn{2}{c}{MIMIC-IV} & \multicolumn{2}{c}{eICU} \\
\cmidrule(lr){2-3}\cmidrule(l){4-5}
 & Training & Test & Training & Test \\
\midrule
Admissions & 305,236 & 38,218 & 116,338 & 14,527 \\
Age at admission (years) & 63 [48, 75] & 63 [49, 75] & 65 [53, 76] & 65 [53, 76] \\
Female, $n$ (\%) & 161,720 (53.0) & 19,707 (51.6) & 53,968 (46.4) & 6,800 (46.8) \\
Length of stay (days) & 4.13 [2.67, 7.08] & 4.16 [2.67, 7.05] & 5.69 [3.33, 9.66] & 5.74 [3.35, 9.75] \\
\midrule
\textit{Test outcomes} & 168 h & 415 h & 168 h & 494 h \\
\midrule
Admissions with measured targets & 29,537 & 29,595 & 13,738 & 13,741 \\
Physiological measurements & 4,706,751 & 6,413,557 & 25,030,956 & 34,169,838 \\
Admissions in mortality comparison & 38,215 & 38,215 & 14,410 & 14,410 \\
Hospital deaths by horizon & 506 & 757 & 635 & 946 \\
\bottomrule
\end{tabular}
\end{table}

\subsection{Clinical inputs and training}
\label{sec:full-variable}

All HealthArc configurations were trained separately within each cohort on the same forecast targets. Within both MEDS conversions, entries carrying medication, infusion, treatment, procedure or HCPCS event-code prefixes were routed deterministically to the intervention stream $\mathcal A_i$, and all other timestamped clinical entries to the observation stream $\mathcal O_i$ (Appendix~\ref{app:language-interface}). For a fair comparison with fixed-channel baselines, we defined a \emph{common input catalogue} of 41 physiological variables and 20 intervention channels with comparable definitions across both cohorts (Table~\ref{tab:channel-catalogue}), restricting inputs to those 61 timestamped clinical channels without separate static demographic entries. Under this catalogue, we trained HealthArc with both timestamp-wise assimilation and temporally grouped assimilation (Section~\ref{sec:latent-filter}). To test whether HealthArc benefits from broader clinical context without changing its architecture, we also trained the temporally grouped model on the \emph{full clinical history}: all structured clinical event types and static baseline entries (such as sex, birth records and, in eICU, ethnicity) available in the admission's MEDS record by the 36-hour cutoff. Alongside numerical and qualitative measurements, this history includes structured bedside assessments, care-plan items, diagnoses timestamped before the cutoff and broader medication and treatment records. Neither study cohort includes longitudinal in-stay free-text progress notes: our MIMIC-IV v3.1 MEDS conversion uses the core structured hospital and ICU tables (its separate notes release provides discharge summaries and radiology reports rather than daily inpatient progress notes), and eICU v2.0 omits unstructured free-text clinical notes altogether for de-identification. Both grouped models used the same encoder architecture and state dimension. Each baseline received the same 61 common-catalogue variables in the input representation required by its architecture (Appendix~\ref{app:baseline-adaptation}).

\subsection{Baselines}
\label{sec:baselines}

For physiological forecasting, we compared HealthArc with three architectures designed for irregular multivariate time series: APN aggregates observations into adaptive temporal patches \citep{liu2026apn}, ASTGI predicts at a query time by aggregating neighboring observations on a spatio-temporal graph \citep{liu2026astgi}, and GRUwE updates a recurrent state using exponential functions of elapsed time \citep{joshi2026gruwe}. For mortality prediction, we compared against three longitudinal survival models: Dynamic-DeepHit applies temporal attention over a recurrent history \citep{lee2020dynamic}, SurvLatent ODE encodes irregular history with an ODE-RNN \citep{moon2022survlatent}, and TrajSurv learns continuous latent trajectories with a neural CDE \citep{zeng2025trajsurv}. Appendix~\ref{app:baseline-adaptation} details baseline input preparation and task adaptation.

\subsection{Evaluation metrics and representation analyses}

For physiological forecasting, we report standardized mean absolute error (sMAE, primary metric) and standardized root mean squared error (sRMSE), standardizing each variable by its training standard deviation in the common catalogue so errors are comparable across clinical units. Matching each cohort's MEDS evaluation unit, we computed each metric separately per measured target within each MIMIC-IV patient or eICU admission, averaged across measured targets within that unit, and then averaged across units. We obtained 95\% confidence intervals from 2,000 bootstrap resamples over evaluation units.

For mortality prediction, we evaluated discrimination using Harrell's concordance ($C_H$), Uno's inverse-probability-of-censoring-weighted concordance ($C_U$) \citep{uno2011cstat} and time-averaged cumulative/dynamic area under the receiver operating characteristic curve (AUC) \citep{bansal2019roc}. We measured overall survival-probability error across the horizon using the integrated Brier score (IBS) for right-censored data \citep{mogensen2012pec}, which integrates squared probability error over time using censoring weights estimated from the training partition. Concordance metrics rank admissions by estimated net risk of hospital death at the forecast horizon $H$. Appendix~\ref{app:scoring} specifies the evaluation grid and numerical integration.

\paragraph{Representation analyses.}
To examine what clinical information the inferred trajectory encodes for downstream tasks not directly used during training, we froze each cohort's timestamp-wise HealthArc model trained on the common catalogue and fitted regularized logistic classifiers to the cutoff state $z_c$ to predict subsequent acute clinical events in both cohorts. In MIMIC-IV, we also grouped patients by their whole latent trajectories and tested whether patients with nearby forecast states had similar measured organ function (Appendix~\ref{app:latent-protocol}).

\section{Results}
\label{sec:results}

\begin{figure}[!htb]
\centering
\includegraphics[width=\linewidth]{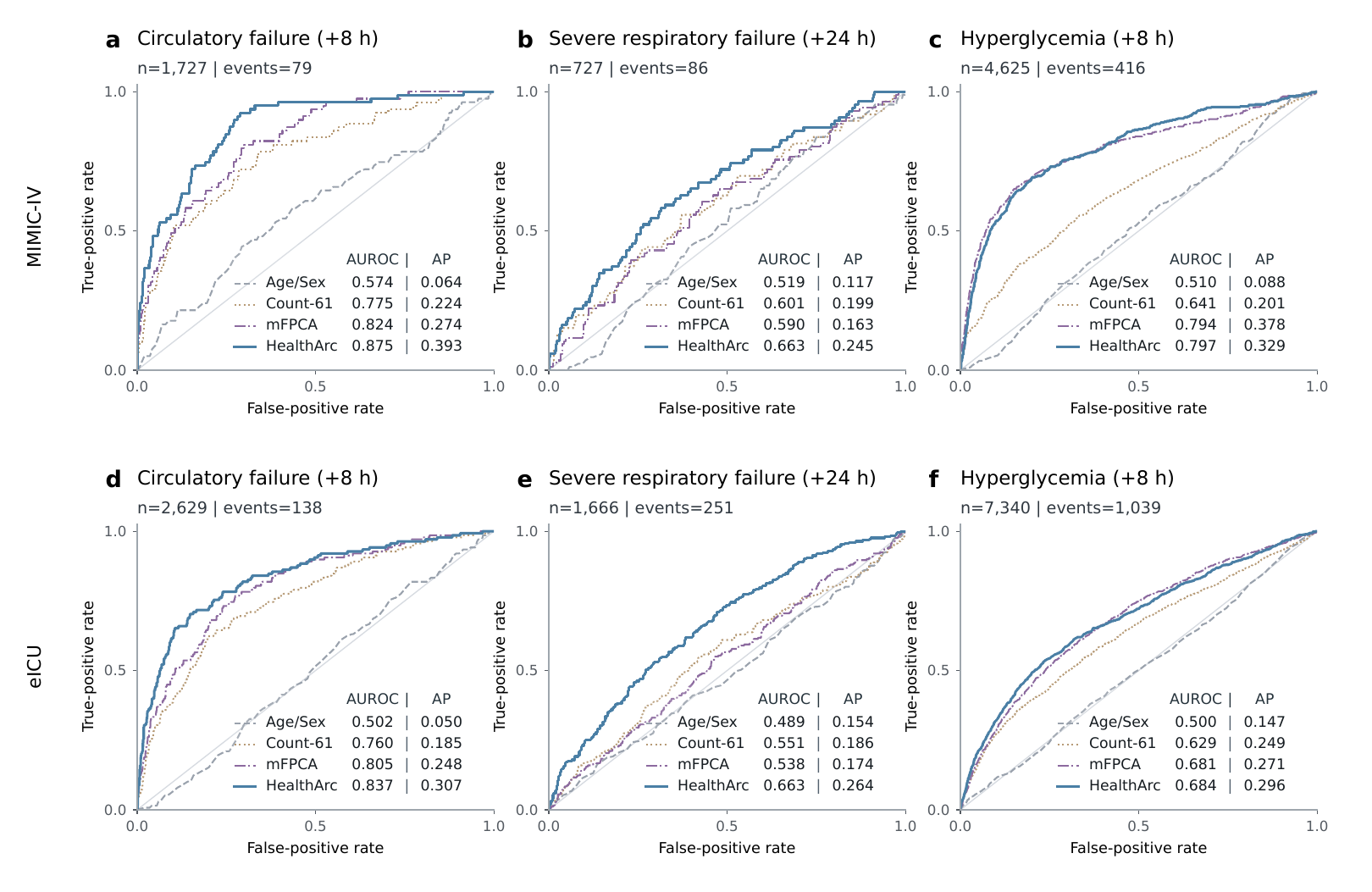}
\caption{\textbf{Clinical event information in the cutoff state.} Regularized logistic classifiers use the frozen cutoff state $z_c$ from HealthArc with timestamp-wise assimilation and the common catalogue. Comparators use age and sex (Age/Sex), per-channel pre-cutoff event counts across the 61 common channels (Count-61) or multivariate functional principal component analysis (mFPCA). Each panel lists the test sample size and positive count above the plot, and area under the receiver operating characteristic curve (AUROC) and average precision (AP) in the legend. Task definitions and fitting procedures appear in Appendix~\ref{app:latent-protocol}.}
\label{fig:event-readout}
\end{figure}

We first report first-week forecasting accuracy across MIMIC-IV and eICU, then examine the clinical structure encoded in cutoff states and full trajectories (Figures~\ref{fig:event-readout} and~\ref{fig:latent-main}), and finally assess update efficiency and multi-week horizons.

\subsection{Physiological and mortality forecasting}

\paragraph{HealthArc reduces first-week physiological error.}
Under the common input catalogue, both HealthArc assimilation schemes achieved lower first-week sMAE and sRMSE than APN, ASTGI and GRUwE in both cohorts (Table~\ref{tab:results-physiology168}). With temporally grouped assimilation, the configuration shared with the full-history evaluation, sMAE reached 0.408 in MIMIC-IV and 0.484 in eICU. Relative to the strongest physiological baseline in each cohort, ASTGI at 0.437 in MIMIC-IV and APN at 0.501 in eICU, these represent sMAE reductions of 6.7\% and 3.4\%, with corresponding sRMSE reductions of 5.5\% and 2.8\%. Comparing the two update schemes, temporally grouped updates gave lower point estimates for both errors in MIMIC-IV, whereas timestamp-wise updates gave lower errors in eICU (sMAE 0.476, sRMSE 0.534).

\begin{table}[t]
\centering
\small
\setlength{\tabcolsep}{2.5pt}
\caption{Physiological forecasting over 168 hours after the 36-hour cutoff. Entries report sMAE and sRMSE averaged over targets and evaluation units ($\downarrow$) with 95\% bootstrap confidence intervals. Bold marks the best point estimate under the common input catalogue. Full clinical history includes all structured clinical event types available in the admission's MEDS record by the cutoff.}
\label{tab:results-physiology168}
\begin{tabular}{@{}lcccc@{}}
\toprule
 & \multicolumn{2}{c}{MIMIC-IV} & \multicolumn{2}{c}{eICU} \\
\cmidrule(lr){2-3}\cmidrule(lr){4-5}
Model & sMAE & sRMSE & sMAE & sRMSE \\
\midrule
APN & \metricci{0.440}{0.438}{0.443} & \metricci{0.493}{0.491}{0.496} & \metricci{0.501}{0.498}{0.503} & \metricci{0.558}{0.555}{0.561} \\
ASTGI & \metricci{0.437}{0.434}{0.439} & \metricci{0.489}{0.486}{0.492} & \metricci{0.530}{0.528}{0.533} & \metricci{0.586}{0.583}{0.589} \\
GRUwE & \metricci{0.444}{0.441}{0.446} & \metricci{0.497}{0.494}{0.499} & \metricci{0.654}{0.651}{0.657} & \metricci{0.709}{0.706}{0.712} \\
\midrule
\multicolumn{5}{l}{\textit{HealthArc, common input catalogue}} \\
\quad Timestamp-wise & \metricci{0.410}{0.407}{0.412} & \metricci{0.464}{0.461}{0.467} & \metricci{\textbf{0.476}}{0.473}{0.478} & \metricci{\textbf{0.534}}{0.531}{0.537} \\
\quad Temporally grouped & \metricci{\textbf{0.408}}{0.405}{0.410} & \metricci{\textbf{0.462}}{0.459}{0.465} & \metricci{0.484}{0.481}{0.487} & \metricci{0.542}{0.539}{0.545} \\
\midrule
\multicolumn{5}{l}{\textit{HealthArc, full clinical history}} \\
\quad Temporally grouped & \metricci{0.406}{0.403}{0.408} & \metricci{0.460}{0.457}{0.463} & \metricci{0.484}{0.481}{0.487} & \metricci{0.542}{0.539}{0.545} \\
\bottomrule
\end{tabular}
\end{table}

\paragraph{HealthArc improves first-week mortality discrimination and probability error.}
Under the same common inputs, both HealthArc assimilation schemes also achieved higher mortality discrimination ($C_H$, $C_U$, AUC) and lower probability error (IBS) than Dynamic-DeepHit, SurvLatent ODE and TrajSurv over the first week (Table~\ref{tab:results-survival168}). With temporally grouped assimilation, HealthArc reached Uno's concordance of 0.814 in MIMIC-IV and 0.747 in eICU, compared with the strongest baseline values of 0.808 for TrajSurv in MIMIC-IV and 0.657 for Dynamic-DeepHit in eICU. Between the two update schemes, timestamp-wise assimilation achieved higher discrimination and lower IBS in MIMIC-IV ($C_U = 0.822$, IBS $0.01272$). In eICU, temporally grouped assimilation achieved higher concordance and AUC ($C_U = 0.747$), whereas timestamp-wise assimilation achieved lower IBS ($0.03171$).

\begin{table}[!htb]
\centering
\small
\setlength{\tabcolsep}{3pt}
\caption{Mortality prediction over 168 hours after the 36-hour cutoff. $C_H$ and $C_U$ denote Harrell's and Uno's concordance, AUC denotes time-averaged cumulative/dynamic AUC, and IBS denotes integrated Brier score. Bold marks the best point estimate under the common input catalogue.}
\label{tab:results-survival168}
\begin{tabular}{lrrrrrrrr}
\toprule
 & \multicolumn{4}{c}{MIMIC-IV} & \multicolumn{4}{c}{eICU} \\
\cmidrule(lr){2-5}\cmidrule(lr){6-9}
Model & $C_H\uparrow$ & $C_U\uparrow$ & AUC $\uparrow$ & IBS $\downarrow$ & $C_H\uparrow$ & $C_U\uparrow$ & AUC $\uparrow$ & IBS $\downarrow$ \\
\midrule
Dynamic-DeepHit & 0.837 & 0.794 & 0.824 & 0.01321 & 0.703 & 0.657 & 0.698 & 0.03337 \\
SurvLatent ODE & 0.814 & 0.769 & 0.802 & 0.01315 & 0.514 & 0.505 & 0.519 & 0.03381 \\
TrajSurv & 0.847 & 0.808 & 0.834 & 0.01321 & 0.615 & 0.582 & 0.626 & 0.03552 \\
\midrule
\multicolumn{9}{l}{\textit{HealthArc, common input catalogue}} \\
\quad Timestamp-wise & \textbf{0.865} & \textbf{0.822} & \textbf{0.853} & \textbf{0.01272} & 0.760 & 0.713 & 0.762 & \textbf{0.03171} \\
\quad Temporally grouped & 0.860 & 0.814 & 0.850 & 0.01296 & \textbf{0.790} & \textbf{0.747} & \textbf{0.792} & 0.03252 \\
\midrule
\multicolumn{9}{l}{\textit{HealthArc, full clinical history}} \\
\quad Temporally grouped & 0.900 & 0.859 & 0.891 & 0.01195 & 0.834 & 0.783 & 0.839 & 0.03058 \\
\bottomrule
\end{tabular}
\end{table}

\paragraph{Full structured histories substantially improve mortality prediction.}
Providing the temporally grouped model with the full structured clinical history instead of the common 61-channel catalogue substantially improved mortality prediction in both cohorts. Uno's concordance rose from 0.814 to 0.859 in MIMIC-IV and from 0.747 to 0.783 in eICU, accompanied by higher time-averaged AUC and lower IBS in both cohorts (Table~\ref{tab:results-survival168}). First-week physiological sMAE fell slightly from 0.408 to 0.406 in MIMIC-IV and remained 0.484 (to three decimal places) in eICU (Table~\ref{tab:results-physiology168}). This pattern is consistent with the two input scopes: the common catalogue already covers the 41 forecast physiological variables, whereas the full structured history additionally supplies static baseline attributes, pre-cutoff diagnoses, care-plan items, bedside assessments and broader medication and treatment records that can carry additional prognostic signal for mortality without changing the model architecture.

\subsection{Clinical structure of learned latent trajectories}
\label{sec:latent-results}

\paragraph{Cutoff states distinguish subsequent acute clinical deterioration.}
To test whether the inferred states transfer to acute clinical event classification tasks not directly used during training, we fitted regularized logistic classifiers to the frozen cutoff state $z_c$ in both cohorts (Figure~\ref{fig:event-readout}). Across all six dataset--task pairs, cutoff-state classifiers distinguished admissions with subsequent circulatory failure, severe respiratory failure or measured hyperglycemia from admissions with confirmed negative assessments, achieving higher area under the receiver operating characteristic curve (AUROC) than comparators based on age and sex, 61-channel event counts (Count-61) or multivariate functional principal component analysis (mFPCA). For circulatory failure, AUROC reached 0.875 in MIMIC-IV and 0.837 in eICU. Average precision (AP) was also highest for HealthArc on five of the six tasks, whereas mFPCA achieved higher AP on MIMIC-IV hyperglycemia (0.378 versus 0.329). Table~\ref{tab:event-probes} reports 95\% bootstrap confidence intervals for HealthArc's AUROC and AP.

\begin{figure}[!htb]
\centering
\includegraphics[width=\linewidth]{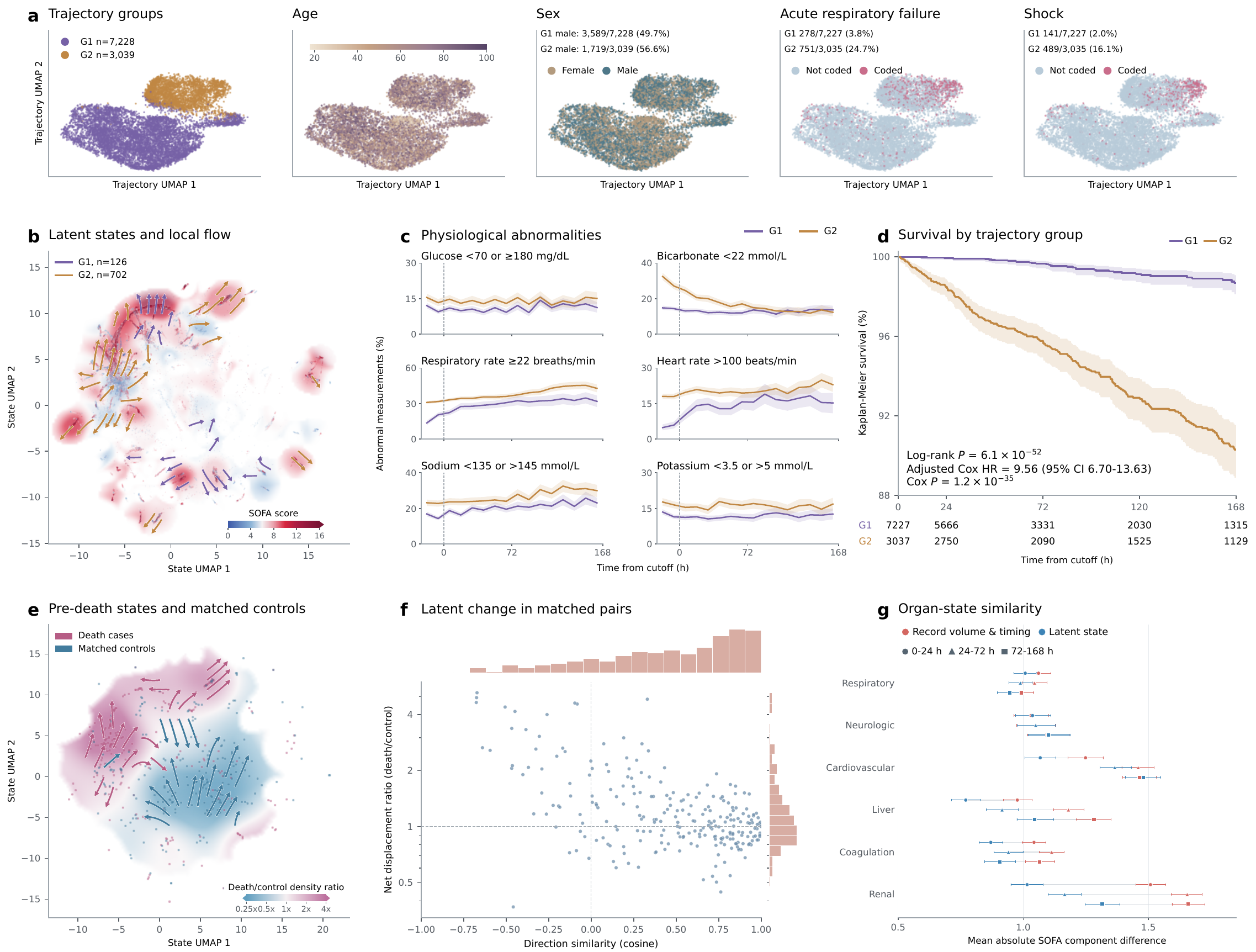}
\caption{\textbf{Clinical organization of HealthArc trajectories.} All panels show analyses in the MIMIC-IV test cohort. (a) Whole trajectories embedded with UMAP and colored by unsupervised trajectory group (G1, G2) and admission characteristics. (b) Individual states at complete six-component SOFA assessments, colored by the patient-balanced local mean SOFA score. (c) Patient-averaged fractions of abnormal measurements in successive 12-hour windows, with pointwise 95\% bootstrap intervals. (d) Kaplan--Meier survival curves (95\% confidence bands) censoring live discharges. The displayed Cox hazard ratio is adjusted for age and sex across all 10,264 patients with valid follow-up. (e) Forecast states for age- and sex-matched case--control pairs evaluated at the same elapsed time since cutoff before the case's death. Pink and blue shading indicate higher relative density of death cases and controls, respectively. Arrows in (b) and (e) show local directions of state change where split-half patient samples agree, with darker arrows indicating greater patient support. (f) Direction cosine and displacement-length ratio for paired displacements from cutoff, $z^{\mathrm{pred}}(\tau)-z^{\mathrm{pred}}(0)$, across the 244 pairs with $\tau > 0$. (g) Mean absolute differences in measured SOFA components for training neighbors selected by latent state versus recording volume and timing. Symbols mark the three forecast periods and whiskers show pointwise 95\% patient-bootstrap intervals.}
\label{fig:latent-main}
\end{figure}

\paragraph{Unsupervised trajectories separate clinical courses and hospital survival.}
Moving from a single cutoff snapshot to full trajectories in MIMIC-IV, unsupervised grouping of trajectories separated hospital courses with sharply different clinical profiles and survival. Using dynamic time warping (DTW) on training trajectories from 24 hours before cutoff through the 168-hour forecast week, without clinical scores or outcome labels, we fitted two reference paths and assigned each of the 10,267 eligible test patients to its closer path (7,228 in group G1 and 3,039 in group G2). Admission-level diagnosis codes for acute respiratory failure (24.7\% versus 3.8\%) and shock (16.1\% versus 2.0\%) were much more frequent in G2 than in G1 (Figure~\ref{fig:latent-main}a).

Mapping individual states at complete six-component Sequential Organ Failure Assessment (SOFA) evaluations showed higher local mean scores along the upper arc of the Uniform Manifold Approximation and Projection (UMAP) embedding and in smaller peripheral regions (Figure~\ref{fig:latent-main}b). Even across regions of comparably high SOFA scores, model-implied transition arrows pointed in different directions, indicating divergent paths of latent-state evolution. Complete six-component SOFA assessments occurred more frequently in G2, which contributed 702 of the 828 patients on this map.

Across successive 12-hour windows (Figure~\ref{fig:latent-main}c), low bicarbonate was initially more common in G2 before the two groups converged later in the week, while elevated respiratory and heart rates together with abnormal glucose, sodium and potassium remained more frequent in G2 throughout. Patient-averaged abnormal respiratory-rate fractions over the forecast week were 34.1\% in G2 versus 27.4\% in G1 among assessed patients (Appendix~\ref{app:latent-protocol}).

Over the first week, 36 of 7,227 G1 patients with mortality follow-up died in hospital, compared with 204 of 3,037 in G2, yielding an age- and sex-adjusted Cox hazard ratio of 9.56 (95\% CI 6.70--13.63; Figure~\ref{fig:latent-main}d). Since sicker patients are also monitored more intensively, we also fitted a sensitivity Cox model among patients with complete pre-cutoff recording features (timestamp count, total text length, time since last observation and median inter-observation gap), where the G2-to-G1 hazard ratio remained 4.11 (95\% CI 2.32--7.30). Whole-path membership thus retained substantial prognostic association beyond charting volume and timing.

\paragraph{Pre-death forecast states occupy shifted latent regions.}
Independent of the two trajectory groups, pre-death forecast states concentrated in distinct regions of latent space relative to age- and sex-matched controls. We compared MIMIC-IV test patients who died in hospital with matched controls still under follow-up at each case's death time, evaluating both at the last forecast time before death (the same elapsed time since cutoff). Although the two distributions overlapped in UMAP projection, pre-death states concentrated along the upper and left regions whereas controls concentrated toward the lower right (Figure~\ref{fig:latent-main}e).

To inspect post-cutoff movement in the original 256-dimensional latent space, we computed each patient's displacement from cutoff, $z^{\mathrm{pred}}(\tau) - z^{\mathrm{pred}}(0)$, across the 244 pairs with $\tau > 0$. Paired displacement vectors formed an acute angle (positive cosine similarity) in 203 of 244 pairs, with a median cosine of 0.543 (interquartile range 0.180--0.819) and a median case-to-control displacement length ratio of 1.069 (interquartile range 0.883--1.456; Figure~\ref{fig:latent-main}f). Thus, paired pre-death and control trajectories occupied shifted regions in the projected state map while showing partial directional alignment, rather than opposite movement, in their post-cutoff net displacements.

\paragraph{Latent-state neighbors have more similar measured organ function.}
To test directly whether proximity in forecast state reflects shared organ physiology rather than shared charting patterns, we analyzed MIMIC-IV test patients with complete SOFA assessments during follow-up. For each query assessment, we assembled a candidate pool of training patients matched on elapsed time since cutoff, historical timestamp count and total text length, and compared two sets of five nearest neighbors drawn from that same pool: one selected by Euclidean distance in the 256-dimensional latent space and a comparator selected by Euclidean distance on standardized pre-cutoff recording features (Appendix~\ref{app:latent-protocol}).

Summing absolute differences across all six measured SOFA components and averaging equally across patients, latent-state neighbors had a mean six-component SOFA distance of 6.159 versus 7.146 for recording-feature neighbors, a 13.8\% reduction across the forecast horizon (paired patient-bootstrap 95\% CI 12.3--15.3\%; Figure~\ref{fig:latent-main}g). Renal, liver and coagulation components showed the largest reductions, whereas the confidence interval for the neurologic component included zero.

\subsection{Update frequency and forecast horizon}

\paragraph{Temporal grouping reduces dense update counts and batch time.}
Within the first 36 hours since admission, common-catalogue entries arrived at an average of 317.1 distinct timestamps per test admission in eICU versus 13.2 in MIMIC-IV. Across the 14,527 eICU test admissions, temporally grouped assimilation of the full clinical history reduced the number of state-update times from 5,337,157 to 821,460 (84.6\%) while preserving all intervention timestamps. In a controlled timing benchmark on identical eICU minibatches with the common inputs, temporal grouping reduced median iteration time across input preparation, forward pass and backpropagation on six GPUs from 29.46 to 25.80 seconds per minibatch (12.4\%; Appendix~\ref{app:scoring}).

\paragraph{Physiological gains persist over longer horizons, whereas survival probability error degrades.}
Forecasting to the 95th-percentile follow-up horizons (415 hours in MIMIC-IV and 494 hours in eICU) from the same 36-hour history, both HealthArc assimilation schemes maintained lower sMAE and sRMSE than all three physiological baselines in both cohorts (Appendix~\ref{app:long-horizon-results}). For mortality prediction under common inputs, temporally grouped HealthArc had lower concordance than TrajSurv in MIMIC-IV ($C_U = 0.677$ versus $0.712$) and higher IBS than SurvLatent ODE in eICU ($0.09248$ versus $0.08863$). At these extended horizons, all three HealthArc configurations had higher IBS than the cohort-level Kaplan--Meier reference, giving negative Brier skill scores (Table~\ref{tab:brier-skill}). Physiological forecasts from the initial 36-hour window thus remained stronger than all baselines over multi-week horizons, whereas survival probability error from the fixed 36-hour history exceeded the cohort Kaplan--Meier reference (Section~\ref{sec:limitations}).

\FloatBarrier
\section{Discussion}
\label{sec:limitations}

Longitudinal hospital records are irregular, sparse and heterogeneous, leading many conventional clinical time-series models to depend on fixed variable panels or explicit missingness handling. HealthArc addresses this mismatch through language-informed data assimilation: observations enter a shared natural-language embedding space and revise the continuous-time latent state through prediction residuals, so even a single measurement or bedside assessment can update the trajectory, and expanding from 61 common channels to the full structured MEDS history required no architectural changes. Across MIMIC-IV and eICU, this unified architecture outperformed task-specific physiological and survival baselines over the first week under matched inputs, supported cutoff-state discrimination of acute clinical deterioration, and delivered substantial additional mortality gains when broader structured context became available.

A central interpretive question in longitudinal EHR modeling is whether learned representations track patient physiology or mainly mirror documentation intensity, since testing frequency and timing reflect clinical decisions \citep{hripcsak2013next,agniel2018biases}. In MIMIC-IV, the two unsupervised whole-trajectory groups differed sharply in admission-level diagnosis codes for acute respiratory failure and shock as well as hospital survival, and complete SOFA measurements were concentrated in the higher-risk group. Two controlled analyses further examine these patterns after accounting for measured pre-cutoff recording features. First, the G2-to-G1 mortality association remained substantial after adjusting for pre-cutoff recording volume and timing (hazard ratio 4.11). Second, among patients matched on forecast time and pre-cutoff recording volume, latent-state neighbors had 13.8\% smaller measured six-component SOFA differences than neighbors selected by recording features alone, indicating closer alignment in measured organ function (Figure~\ref{fig:latent-main}g).

Our cross-cohort evaluation also clarifies how data sparsity interacts with update granularity. With the common catalogue, eICU test admissions averaged about 24 times as many distinct recording times in the first 36 hours as MIMIC-IV test admissions. An individual laboratory variable can still go unmeasured for hours while other observations update the state. When charting is dense, temporally grouped assimilation offers a practical middle ground: it preserves chronological order and relative timestamps within one-hour windows and keeps intervention times exact, cutting eICU update counts by 84.6\% and batch time by 12.4\%. Accuracy trade-offs depended on both task and cohort, with grouping improving first-week eICU mortality concordance and MIMIC-IV physiological error, whereas timestamp-wise updates retained lower eICU physiological error and stronger MIMIC-IV mortality metrics.

Several modeling boundaries define where these results hold and where further work is needed. Over multi-week horizons (415\,h and 494\,h), HealthArc's IBS exceeded the cohort-level Kaplan--Meier reference (Table~\ref{tab:brier-skill}). Although post-cutoff measurements and outcomes supervise training, inference uses only the first 36 hours without subsequent assimilation. Intervening clinical events and treatment changes likely weaken the signal from that initial window, a hypothesis that rolling post-cutoff updates can test directly. Reported mortality risks also condition on a single posterior-mean initial state without Monte Carlo uncertainty propagation and treat live discharge as right censoring. Sampling from $q_\psi(\Delta \mid z_c)$ at evaluation and modeling live discharge as a competing risk would enable uncertainty intervals and direct estimation of cumulative mortality incidence before discharge.

Additional boundaries stem from data modalities and care settings. Our partition of observations and interventions relied on structured event-code prefixes from relational tables, as neither study cohort includes timestamped in-stay free-text progress notes. In prospective clinical deployments, however, narrative admission, nursing and daily progress notes arrive at intermediate timestamps throughout a hospital stay alongside structured measurements. Within such free-text entries, clinical observations and treatment plans often appear in the same sentence or paragraph, so extending HealthArc to prospective narrative streams will require either an upstream extraction step or a learned routing mechanism. Both evaluation cohorts also consist of inpatient admissions, leaving HealthArc's behavior across sparser multi-year outpatient records to be tested.

Together, these results show that pairing continuous-time latent dynamics with a shared clinical language interface supports accurate forecasting and clinically structured state estimation from sparse, heterogeneous inpatient records without rigid input schemas. Extending this assimilation framework to rolling post-cutoff updates, narrative clinical text and longer-term care trajectories is a natural next step toward general continuous-time patient models.

\bibliographystyle{plainnat}
\bibliography{reference}

\appendix
\section{Model and training details}
\label{app:implementation}

\paragraph{Language encoding, event routing and temporal grouping.}
\label{app:language-interface}
We kept the base weights of Qwen3-Embedding-4B \citep{zhang2025qwen3} frozen and applied low-rank adaptation (LoRA) \citep{hu2022lora} with rank 8 in the query, key, value and output attention projections. For structured MEDS records in both MIMIC-IV and eICU, entries from medication, pharmacy, infusion, procedure and treatment tables whose event codes begin with \texttt{MEDICATION//}, \texttt{MEDICATION\_END//}, \texttt{INFUSION}, \texttt{TREATMENT//}, \texttt{PROCEDURE//} or \texttt{HCPCS//} are routed to the intervention stream $\mathcal A_i$. All other timestamped clinical entries are routed to the observation stream $\mathcal O_i$. In the full clinical history configuration, static baseline entries (untimestamped demographic attributes and eligible \texttt{MEDS\_BIRTH} records) are serialized and encoded via $E_\eta$. The static projection input $e_{\mathrm{stat}} \in \R^{2d_e}$ concatenates pooled static observation and static intervention embeddings (zero-filled when absent) and is mapped by a two-layer multilayer perceptron (MLP) $P_{\mathrm{stat}}$ to the 192-dimensional static context $\xi$. Under the common 61-channel catalogue, only matched timestamped clinical channels are retained and both static branches receive zero embeddings, so the same MLP outputs a shared default static context. When serialized observations at a single timestamp exceed the character limit per segment, we split them at event boundaries and pool the normalized segment embeddings with weights proportional to their token counts. For temporally grouped assimilation, we set the maximum group span to one hour, grouping consecutive observation-only timestamps as long as the elapsed time from the group's first timestamp stays within this limit and the serialized text fits within the encoder's context window, with each entry carrying its time offset relative to the group's final timestamp. Any timestamp containing an intervention closes the preceding group and is processed at its exact recorded time together with any simultaneous observations.

\paragraph{Dynamics and training.}
The latent state $z(t)$ has dimension $d_z = 256$, the intervention control $u_i$ has dimension $d_u = 192$, and the static context $\xi$ has dimension $d_\xi = 192$. The predictor $G_\omega$, static projection $P_{\mathrm{stat}}$, state initializer $P_{\mathrm{init}}$, intervention projection $P_a$ and state-correction network $F_\phi$ are implemented as MLPs with SiLU activations. In Equation~\eqref{eq:observation-update}, $\mathrm{LN}_a$ and $\mathrm{LN}_z$ are Layer Normalization layers with learned elementwise scale and bias, whereas $\mathrm{LN}_0$ in Section~\ref{sec:historical-semantics} is parameter-free Layer Normalization. The multiplicative gate $\tanh(W_\phi r_i)$ uses a bias-free linear projection $W_\phi$ of the prediction residual $r_i$. Each continuous-time interval ending at an update, cutoff or query time is integrated using equal fourth-order Runge--Kutta (RK4) steps of at most one hour. Both the historical embedding loss $\mathcal L_{\rm hist}$ and physiological loss $\mathcal L_{\rm phys}$ use the smooth L1 (Huber) penalty $\rho$ with a transition threshold of $1$. In Equation~\eqref{eq:total-loss}, we set $\lambda_{\rm mort} = 1$, $\lambda_{\rm kin} = 10^{-3}$, $\lambda_{\rm hist} = 0.1$ and $\beta_{\rm cut} = 10^{-2}$, with an isotropic cutoff prior standard deviation $\sigma_0 = 0.1$ per coordinate. Each model was trained for one epoch with AdamW. The mortality likelihood used exact observed death or censoring times, and reported survival probabilities were evaluated without post-hoc calibration.

\paragraph{Physiological prediction head.}
For each physiological variable, the 1st and 99th percentiles of the training measurements define clipping bounds. Future training targets and the last pre-cutoff measurement $b_v$ in Equation~\eqref{eq:phys-anchor} are clipped to these bounds and standardized using the mean and standard deviation of the clipped training values.

\subsection{Baseline input representations}
\label{app:baseline-adaptation}

Pre-cutoff clinical information from the common catalogue was formatted according to each baseline architecture. APN, ASTGI and GRUwE received irregular numerical values, observation masks and timestamps, retaining repeated measurements as separate sparse observations. SurvLatent ODE used hourly aggregation. Dynamic-DeepHit and SurvLatent ODE were adapted to the single in-hospital death endpoint using their respective discrete-time and latent-ODE survival parameterizations. TrajSurv received pre-cutoff SOFA scores for its auxiliary contrastive supervision.

\paragraph{Common input catalogue.}
We selected 41 physiological variables with comparable definitions and units across both MIMIC-IV and eICU, spanning vital signs, serum chemistries, liver tests, blood counts and coagulation panels, together with 20 intervention channels covering hemodynamic support, sedation/analgesia, other major ICU medications, ventilatory support and renal replacement therapy (Table~\ref{tab:channel-catalogue}). Every baseline and common-input HealthArc model received the same 61 clinical variables in its respective input format.

\begin{table}[!htb]
\centering
\small
\caption{Physiological targets and intervention inputs in the common catalogue.}
\label{tab:channel-catalogue}
\begin{tabular}{@{}p{0.24\linewidth}p{0.70\linewidth}@{}}
\toprule
Clinical group & Channels \\
\midrule
\multicolumn{2}{@{}l}{\textit{Physiological measurements (41)}} \\
Vital signs & Heart rate, respiratory rate, oxygen saturation, systolic and diastolic blood pressure \\
Chemistry & Glucose, creatinine, blood urea nitrogen, sodium, potassium, chloride, bicarbonate (or total CO$_2$), anion gap, total calcium, magnesium, phosphate, lactate \\
Proteins and liver tests & Albumin, total protein, total and direct bilirubin, aspartate aminotransferase, alanine aminotransferase, alkaline phosphatase \\
Blood counts and red cell indices & Hemoglobin, hematocrit, platelet count, white and red blood cell counts, mean corpuscular hemoglobin, mean corpuscular volume, mean corpuscular hemoglobin concentration, red cell distribution width \\
Differential counts & Neutrophils, lymphocytes, monocytes, eosinophils and basophils, each as a percentage of white blood cells \\
Coagulation & Prothrombin time, partial thromboplastin time, fibrinogen \\
\midrule
\multicolumn{2}{@{}l}{\textit{Intervention families (20)}} \\
Hemodynamic support & Norepinephrine, epinephrine, phenylephrine, vasopressin, dopamine, dobutamine, milrinone \\
Sedation and analgesia & Propofol, fentanyl, midazolam, dexmedetomidine \\
Other medications & Insulin, furosemide, heparin, vancomycin, piperacillin/tazobactam \\
Respiratory and renal support & Invasive or unspecified mechanical ventilation, noninvasive ventilation, hemodialysis, continuous renal replacement therapy \\
\bottomrule
\end{tabular}
\end{table}

\subsection{Scoring and statistical reporting}
\label{app:scoring}

The longer forecast horizons were derived from training follow-up durations. In MIMIC-IV, the 95th percentile of remaining stay after the 36-hour cutoff among eligible training admissions was 414.49 hours. In eICU, the 95th percentile of total stay among source training admissions was 529.69 hours, or 493.69 hours after subtracting the 36-hour history. Rounding up gave 415 and 494 hours, respectively.

For the computation benchmark, we processed the same four minibatches (180 eICU admissions in total) under each assimilation scheme. Each minibatch was executed three times across six GPUs with identical model initialization and no optimizer updates, timing input preparation, forward propagation and backpropagation. We report the median across the 12 iteration times per scheme, recording the slowest GPU on each iteration.

For mortality scoring, predicted risks were evaluated on the grid $\{6, 12, 24, 36, 48, 72, 96, 120, 144, 168\}$ hours since cutoff, appending the terminal 415-hour or 494-hour time point for longer horizons. Cumulative/dynamic AUC was trapezoidally averaged from 6 hours to $H$. IBS was integrated from $\tau = 0$ with $\mathrm{BS}(0) = 0$. Long-horizon integration retained the first-week grid and added a single trapezoid between 168 hours and the final query time.

\section{Longer-horizon results}
\label{app:long-horizon-results}

Tables~\ref{tab:results-mimic415} and~\ref{tab:results-eicu494} report forecasts to 415 hours in MIMIC-IV and 494 hours in eICU from the same 36-hour pre-cutoff history. Both HealthArc assimilation schemes maintained lower sMAE and sRMSE than all three physiological baselines in both cohorts, with timestamp-wise updates attaining the lowest physiological errors among common-input models. For mortality prediction under common inputs, TrajSurv achieved higher concordance and AUC in MIMIC-IV while Dynamic-DeepHit achieved the lowest IBS, and SurvLatent ODE achieved the lowest IBS in eICU. Expanding input access to the full structured clinical history improved concordance, AUC and IBS for temporally grouped HealthArc in both cohorts. Table~\ref{tab:brier-skill} reports the corresponding integrated Brier skill scores ($\mathrm{BSS} = 1 - \mathrm{IBS}/\mathrm{IBS}_{\rm KM}$) relative to the training-cohort Kaplan--Meier reference.

\begin{table}[!htb]
\centering
\small
\setlength{\tabcolsep}{3pt}
\caption{MIMIC-IV forecasts to 415 hours after the 36-hour cutoff. Physiological entries include 95\% bootstrap confidence intervals. Bold identifies the best point estimate under the common input catalogue. A dash denotes a task not evaluated.}
\label{tab:results-mimic415}
\begin{tabular}{@{}lccrrrr@{}}
\toprule
Model & sMAE $\downarrow$ & sRMSE $\downarrow$ & $C_H\uparrow$ & $C_U\uparrow$ & AUC $\uparrow$ & IBS $\downarrow$ \\
\midrule
APN & \metricci{0.435}{0.433}{0.438} & \metricci{0.492}{0.489}{0.494} & --- & --- & --- & --- \\
ASTGI & \metricci{0.443}{0.440}{0.445} & \metricci{0.497}{0.495}{0.500} & --- & --- & --- & --- \\
GRUwE & \metricci{0.457}{0.455}{0.460} & \metricci{0.513}{0.510}{0.515} & --- & --- & --- & --- \\
Dynamic-DeepHit & --- & --- & 0.819 & 0.700 & 0.753 & \textbf{0.03968} \\
SurvLatent ODE & --- & --- & 0.767 & 0.653 & 0.717 & 0.03971 \\
TrajSurv & --- & --- & \textbf{0.835} & \textbf{0.712} & \textbf{0.769} & 0.04342 \\
\midrule
\multicolumn{7}{l}{\textit{HealthArc, common input catalogue}} \\
\quad Timestamp-wise & \metricci{\textbf{0.415}}{0.412}{0.417} & \metricci{\textbf{0.471}}{0.469}{0.474} & 0.783 & 0.659 & 0.734 & 0.04083 \\
\quad Temporally grouped & \metricci{0.416}{0.414}{0.419} & \metricci{0.472}{0.469}{0.475} & 0.820 & 0.677 & 0.754 & 0.04093 \\
\midrule
\multicolumn{7}{l}{\textit{HealthArc, full clinical history}} \\
\quad Temporally grouped & \metricci{0.413}{0.410}{0.415} & \metricci{0.469}{0.466}{0.472} & 0.841 & 0.687 & 0.774 & 0.04029 \\
\bottomrule
\end{tabular}
\end{table}

\begin{table}[!htb]
\centering
\small
\setlength{\tabcolsep}{3pt}
\caption{eICU forecasts to 494 hours after the 36-hour cutoff. Physiological entries include 95\% bootstrap confidence intervals. Bold marks the best common-input scores, including ties at the reported precision. A dash denotes a task not evaluated.}
\label{tab:results-eicu494}
\begin{tabular}{@{}lccrrrr@{}}
\toprule
Model & sMAE $\downarrow$ & sRMSE $\downarrow$ & $C_H\uparrow$ & $C_U\uparrow$ & AUC $\uparrow$ & IBS $\downarrow$ \\
\midrule
APN & \metricci{0.512}{0.510}{0.515} & \metricci{0.573}{0.570}{0.576} & --- & --- & --- & --- \\
ASTGI & \metricci{0.556}{0.553}{0.558} & \metricci{0.615}{0.612}{0.618} & --- & --- & --- & --- \\
GRUwE & \metricci{0.651}{0.648}{0.654} & \metricci{0.710}{0.707}{0.713} & --- & --- & --- & --- \\
Dynamic-DeepHit & --- & --- & 0.680 & 0.614 & 0.618 & 0.09446 \\
SurvLatent ODE & --- & --- & 0.596 & 0.563 & 0.583 & \textbf{0.08863} \\
TrajSurv & --- & --- & 0.599 & 0.540 & 0.573 & 0.10680 \\
\midrule
\multicolumn{7}{l}{\textit{HealthArc, common input catalogue}} \\
\quad Timestamp-wise & \metricci{\textbf{0.489}}{0.487}{0.492} & \metricci{\textbf{0.551}}{0.548}{0.555} & 0.704 & \textbf{0.618} & 0.641 & 0.09846 \\
\quad Temporally grouped & \metricci{0.492}{0.489}{0.495} & \metricci{0.554}{0.551}{0.558} & \textbf{0.711} & \textbf{0.618} & \textbf{0.668} & 0.09248 \\
\midrule
\multicolumn{7}{l}{\textit{HealthArc, full clinical history}} \\
\quad Temporally grouped & \metricci{0.493}{0.490}{0.496} & \metricci{0.555}{0.551}{0.558} & 0.798 & 0.683 & 0.724 & 0.09186 \\
\bottomrule
\end{tabular}
\end{table}

\begin{table}[!htb]
\centering
\small
\caption{Integrated Brier skill relative to the training-set Kaplan--Meier reference, $\mathrm{BSS}=1-\mathrm{IBS}/\mathrm{IBS}_{\rm KM}$. Positive values indicate lower probability error than this reference. The reference IBS was 0.01350 and 0.04019 at 168 and 415 hours in MIMIC-IV, and 0.03386 and 0.08899 at 168 and 494 hours in eICU.}
\label{tab:brier-skill}
\begin{tabular}{lrrrr}
\toprule
Model & MIMIC 168h & MIMIC 415h & eICU 168h & eICU 494h \\
\midrule
Dynamic-DeepHit & 0.022 & 0.013 & 0.015 & -0.061 \\
SurvLatent ODE & 0.026 & 0.012 & 0.002 & 0.004 \\
TrajSurv & 0.022 & -0.080 & -0.049 & -0.200 \\
\midrule
\multicolumn{5}{l}{\textit{HealthArc, common input catalogue}} \\
\quad Timestamp-wise & 0.058 & -0.016 & 0.064 & -0.106 \\
\quad Temporally grouped & 0.040 & -0.018 & 0.040 & -0.039 \\
\midrule
\multicolumn{5}{l}{\textit{HealthArc, full clinical history}} \\
\quad Temporally grouped & 0.115 & -0.002 & 0.097 & -0.032 \\
\bottomrule
\end{tabular}
\end{table}

\section{Representation and trajectory analyses}
\label{app:latent-protocol}

\subsection{Analysis units and source cohorts}

The trajectory analyses in Figure~\ref{fig:latent-main} used the MIMIC-IV model, while the clinical event classifiers in Figure~\ref{fig:event-readout} used each cohort's own model. All models in these analyses were trained with timestamp-wise assimilation and the common 61-channel catalogue. Event classifiers took the deterministic cutoff state $z_c$ as input, and forecast trajectories started from the posterior mean $z^{\mathrm{pred}}(0) = z_c + \mu_\Delta(z_c)$ (Equation~\eqref{eq:cutoff-latent}).

Within MIMIC-IV, the whole-path analysis (Figure~\ref{fig:latent-main}a--d) required recorded observations at least 24 hours before the cutoff and at least two pre-cutoff updates, yielding 64,505 training, 16,169 validation and 10,267 test patients. States were evaluated every three hours from 24 hours before the cutoff through 168 hours after it. For each historical query time, we propagated the most recent updated state to that query time under its active intervention control and applied the learned posterior-mean offset $z \mapsto z + \mu_\Delta(z)$, placing historical and forecast states in a shared coordinate frame that joins continuously at $z^{\mathrm{pred}}(0)$.

We fitted two reference trajectories to the training paths via DTW and DTW barycenter averaging \citep{petitjean2011dba} in the original 256-dimensional latent space, aligning the pre-cutoff and post-cutoff segments separately at the cutoff. For visualization, UMAP \citep{mcinnes2018umap} embedded whole trajectories by DTW distance and individual states by their principal components. Local flow arrows summarize adjacent state transitions in regions where independent patient half-samples agreed on direction.

\subsection{Clinical descriptions and prognostic association}

SOFA scores followed the six-component system \citep{vincent1996sofa} evaluated in rolling 24-hour windows. Complete six-component scores were available for 828 MIMIC-IV test patients (126 of 7,228 in G1 and 702 of 3,039 in G2), contributing 39,629 assessment states to Figure~\ref{fig:latent-main}b. Panel (c) tracks six physiological abnormality curves (glucose, bicarbonate, respiratory rate, heart rate, sodium and potassium). Within each 12-hour window, we computed the fraction of abnormal measurements per assessed patient and averaged equally across patients, omitting windows with fewer than 20 assessed patients and computing pointwise 95\% confidence bands by patient bootstrap.

The primary survival analysis included 10,264 eligible patients with valid follow-up. We computed 95\% confidence bands for Kaplan--Meier estimates using Greenwood standard errors, compared groups with the log-rank test, and estimated the age- and sex-adjusted hazard ratio using a Cox proportional hazards model. In a sensitivity analysis among patients with complete values for all four pre-cutoff recording features, we additionally adjusted the Cox model for pre-cutoff timestamp count, total recorded text length, time since the last observation and median inter-observation gap.

From the 17,369 unique patients in the MIMIC-IV test split, we matched each eligible patient who died in hospital (with valid pre-death evaluation and recording coverage) to one control of the same sex and within five years of age with no recorded in-hospital death during the first post-cutoff week and with both clinical records and mortality follow-up extending to the case's death time, using each control at most once, yielding 245 matched pairs. At the last forecast evaluation time before each death, we compared case and control states at the identical elapsed time since cutoff. Panel (e) applies a common kernel bandwidth to both projected densities across all 245 pairs, and panel (f) summarizes the direction cosine and norm ratio of paired displacements $z^{\mathrm{pred}}(\tau) - z^{\mathrm{pred}}(0)$ in the original 256-dimensional latent space across the 244 pairs with $\tau > 0$ (excluding one pair whose pre-death evaluation fell at $\tau = 0$).

\subsection{Organ-state neighborhood analysis}

Of the 828 test patients with complete SOFA assessments in panel (b), 657 had forecast assessments with eligible training neighbors for panel (g), yielding 29,213 eligible query assessments matched against 3,800 training reference patients. For each query assessment, we identified eligible training patients who had a complete SOFA assessment within one hour of the query's elapsed time since cutoff (using each training patient's single assessment closest in time) and who had similar pre-cutoff recording volume. Specifically, letting $n$ denote pre-cutoff timestamp count and $L$ denote total serialized text length, a training patient was matched to the query patient if both $(n_{\rm cand} + 1)/(n_{\rm query} + 1)$ and $(L_{\rm cand} + 1)/(L_{\rm query} + 1)$ lay between $1/2$ and $2$, adding $1$ to avoid division by zero. Among queries with at least ten eligible candidates, we selected the five nearest neighbors by Euclidean distance in the original 256-dimensional latent space. As a comparator, we selected five neighbors from the same candidate pool using Euclidean distance on recording features standardized using training statistics (timestamp count, text length, time since last observation and median observation gap) together with their missingness indicators. Measured SOFA scores were held out during neighbor selection and used solely to evaluate clinical similarity afterward.

For each query, we summed absolute differences across all six SOFA components, averaged over the five selected neighbors, averaged within six-hour windows and then averaged within each patient so that every patient contributed equal weight. Paired 95\% intervals used 2,000 patient-bootstrap resamples. Across eight sensitivity analyses varying the primary matching protocol (time tolerance $\pm 0.5$\,h or $\pm 3$\,h; recording-volume ratio bound $1.25$ or $4$, or matching on forecast time alone; $k = 1$ or $k = 10$ neighbors; or cosine instead of Euclidean latent distance), latent-state neighbors reduced six-component SOFA distance relative to recording-feature neighbors by 13.1\% to 15.5\%.

\subsection{Clinical event classification}

\begin{table}[!htb]
\centering
\scriptsize
\setlength{\tabcolsep}{3pt}
\caption{Clinical event classification from the cutoff state $z_c$ of HealthArc with timestamp-wise assimilation. Brackets give pointwise 95\% bootstrap confidence intervals.}
\label{tab:event-probes}
\begin{tabular}{llrrll}
\toprule
Dataset & Task & Test admissions & Positive & AUROC [95\% CI] & AP [95\% CI] \\
\midrule
MIMIC-IV & Circulatory failure (8h) & 1,727 & 79 & 0.875 [0.836, 0.911] & 0.393 [0.286, 0.500] \\
MIMIC-IV & Severe respiratory failure (24h) & 727 & 86 & 0.663 [0.595, 0.721] & 0.245 [0.178, 0.337] \\
MIMIC-IV & Measured hyperglycemia (8h) & 4,625 & 416 & 0.797 [0.772, 0.822] & 0.329 [0.288, 0.375] \\
eICU & Circulatory failure (8h) & 2,629 & 138 & 0.837 [0.797, 0.872] & 0.307 [0.242, 0.391] \\
eICU & Severe respiratory failure (24h) & 1,666 & 251 & 0.663 [0.629, 0.696] & 0.264 [0.225, 0.317] \\
eICU & Measured hyperglycemia (8h) & 7,340 & 1,039 & 0.684 [0.666, 0.701] & 0.296 [0.270, 0.322] \\
\bottomrule
\end{tabular}
\end{table}

We adapted three acute clinical event tasks from ICareFM \citep{burger2026icarefm}: circulatory failure (positive when mean arterial pressure $< 65$\,mmHg and lactate $> 2$\,mmol/L within 8 hours; confirmed negative when mean arterial pressure $> 65$\,mmHg and lactate $< 2$\,mmol/L), severe respiratory failure (positive when measured arterial oxygen partial pressure to inspired oxygen fraction ratio, PaO$_2$/FiO$_2 < 100$\,mmHg within 24 hours; confirmed negative when $> 100$\,mmHg) and measured hyperglycemia (positive when glucose $> 180$\,mg/dL within 8 hours; confirmed negative when $< 180$\,mg/dL). Patients already meeting the positive criterion at cutoff were excluded, while those without a cutoff assessment were retained. Within each evaluation window, an admission was labeled positive if any assessment prior to censoring met the positive criterion, negative if no positive assessment occurred and at least one assessment met the confirmed negative criterion, and unlabeled otherwise (including boundary equality or partial-component states).

For each task, we fitted an L2-regularized logistic classifier to the deterministic cutoff state $z_c$ using fivefold cross-validation grouped by patient, with imputation and feature scaling fitted strictly inside each training fold. Pointwise 95\% confidence intervals for AUROC and AP used 500 patient-bootstrap resamples.

\end{document}